\documentclass{article}

\usepackage{arxiv}

\usepackage[utf8]{inputenc} 
\usepackage[T1]{fontenc}    
\usepackage{hyperref}       
\usepackage{url}            
\usepackage{booktabs}       
\usepackage{amsfonts}       
\usepackage{nicefrac}       
\usepackage{microtype}      
\usepackage{lipsum}		
\usepackage{graphicx}
\usepackage{natbib}
\usepackage{doi}
\usepackage{arydshln}

\title{{\huge Heroes, Villains, and Victims, and GPT-3} \\ Automated Extraction of Character Roles Without Training Data}

\author{Dominik Stammbach \\
ETH Zurich \\
  \texttt{dominsta@ethz.ch} \\\And
  Maria Antoniak \\
  Cornell University \\
  \texttt{maa343@cornell.edu} \\\And
  Elliott Ash \\
ETH Zurich \\
  \texttt{ashe@ethz.ch}
  }

\renewcommand{\shorttitle}{\textit{arXiv} Template}

\hypersetup{
pdftitle={Heroes, Villains, and Victims, and GPT-3},
pdfsubject={q-bio.NC, q-bio.QM},
pdfauthor={Stammbach, Antoniak, Ash},
pdfkeywords={Character Roles Extraction, Narrative Understanding, GPT-3},
}

\begin{document}
\maketitle

\begin{abstract}
This paper shows how to use large-scale pre-trained language models to extract character roles from narrative texts without training data. Queried with a zero-shot question-answering prompt, GPT-3 can identify the \textit{hero}, \textit{villain}, and \textit{victim} in diverse domains: newspaper articles, movie plot summaries, and political speeches. 

\end{abstract}


\section{Introduction}

What makes a good story? According to some leading theories of narrative \cite{propp1968morphology}, a good story has at least three ingredients: a \textit{hero}, a \textit{villain}, and a \textit{victim}. In a classic recipe, the hero defeats the villain to gain justice for the victim. The cohesiveness of these character roles extends even beyond fiction, with social-science research showing the effectiveness of the hero-villain-victim framework in explaining what motivates consumer behavior and political processes \cite{narrative_policy_framework,clement2017hero,bergstrand2018villains}.

Motivated by the relevance of narratives to culture and society, a literature in natural language processing has arisen to automatically detect characters in texts and annotate their character roles \cite{bamman-etal-2013-learning,jahan-finlayson-2019-character, piper-etal-2021-narrative}. In particular, prior work has used dictionary methods to identify heroes, villains, and victims in texts \cite{NFP_and_NLP}. The previous methods have mixed results, motivating the present research.

\begin{figure}
    \center
    \begin{tabular}{p{15cm}} 
\textit{Who is the villain in the following text?}\\
\\

\textit{Text: In questioning the use of hydraulic fracturing in New York to help produce natural gas, you do not note that the technology has been employed and continuously improved for more than 50 years and that studies by the Environmental Protection Agency and the Ground Water Protection Council have not identified a single instance of groundwater contamination. Wells where fracturing is used are specially constructed to protect drinking water sources. Regulatory oversight is extensive. The fluids  mostly water  that are forced into a well to create pressure to fracture rock are pushed back out by the oil and gas flowing upward for safe processing. Protecting our water supplies is important, as are reductions in greenhouse gas emissions through use of clean-burning natural gas. Banning hydraulic fracturing would be unwarranted and shortsighted, preventing production of large amounts of natural gas that could directly benefit New York consumers for decades and create thousands of good jobs.} \\
\\
\textit{Villain:} \textbf{ The villain in this text is the person who is questioning the use of hydraulic fracturing in New York.} \\ \\  \hdashline \\ 

\textit{Hero:} \textbf{The hero in this text is the person who wrote it, who is trying to convince the reader that hydraulic fracturing is a safe and effective way to produce natural gas} \\ 

\textit{Victim:} \textbf{The New York consumers} \\ 
\end{tabular}
\caption{Zero-Shot Example of the NFP using GPT-3. Prompt in italics, GPT-3 completion in bold (Analogous results for the hero and victim below the dashed line)}
\label{fig:fig_1}
\end{figure}

This paper explores the use of large pre-trained language models for the task of character role labeling. Operationalizing the problem as a Machine Reading Comprehension Task (MRCP), we provide an input document and ask the language model who is the hero (or villain or victim). As illustrated by the prompt in Figure \ref{fig:fig_1}, we directly ask ``Who is the hero'' (or villain or victim). Thus, we can extract character roles from plain-text documents without in-domain training data.

We find that a large pre-trained language model, GPT-3 \cite{gpt3}, is more effective in labeling these character roles than previous methods, across a diverse set of narrative domains. First, we investigate our method on a labeled corpus of newspaper articles about fracking where the three roles --- hero, villain, victim --- are manually annotated. In that dataset, our language-model approach is over twice as accurate as an existing baseline using a dictionary approach. Second, as a more qualitative and transparent evaluation, we report the annotations produced on a selection of Disney movie plot summaries. These results comport well with subjective judgment.

In our third experiment, we apply the method to a corpus of U.S. State of the Union Addresses, 2001-2018. Matching up the character role annotations with the party affiliation of the president, we explore partisan differences in the framing of heroes, victims, and villains. To make this process feasible, we explore clustering of the GPT-3 output, which produces more legible sets of character assignments. These results demonstrate the promise of the method for empirical research in social science and the digital humanities.

These results are of broad interest given the narrative centrality of character archetypes \cite{propp1968morphology}. They are of more specific interest in the literature analyzing narrative framing in news media and policy discourse \cite{narrative_policy_framework, Blair_and_McCormack}. A robust and efficient computational method to extract character roles in text without training data opens up a wide array of research questions to quantitative analysis. Our approach can thus help researchers make progress in the largely under-explored area of quantitative narrative analysis  \cite{shiller_2019}.

\section{Related Work}

This paper adds to the work in NLP on automated extraction of character roles from natural language accounts, and in particular the identification of heroes, villains, and victims. The closest paper is \cite{NFP_and_NLP}, who similarly focus on the detection of heroes, villains, and victims in news articles and provide a dictionary-based approach which we will use as a baseline. 

On the broader problem of extracting stereotypical character roles, prior work has explored a variety of methods, including the detection of personas using annotated data combined with feature engineering and regression \cite{bamman_latent_personas}; parsing and pattern matching tools to identify a consistent set of personas (e.g. doctor, nurse, doula) across testimonials about childbirth and then assess the relative power dynamics \cite{antoniak2019narrative}; annotations of German news and social media sentences for villains and rogues and transformer models to machine-tag these roles \cite{klenner-etal-rouges}; clustering of structural plot information from folktales \cite{jahan-etal-2021-inducing}; and a combination of NER and clustered phrase embeddings to identify repeatedly occurring entities, along with semantic role labeling to identify how entities are connected by actions \cite{ash_narratives}. Our method does not rely on labeled data, but we employ some of these techniques (e.g., clustering) to support the legibility of our results.

The second related literature is treating role extraction as a machine reading comprehension (MRCP) task, which for example has been proposed for semantic role labeling \cite{SRL_as_QA}. Most related to our work, \cite{liu-etal-2020-event} and \cite{du-cardie-2020-event} interpret event extraction as an MRPC task and leverage pre-trained language models to extract events, producing state-of-the-art results in event extraction and leading us to apply this method for detection of character roles.

In principle, any NLP task can be framed as MRCP or question answering (QA) tasks \cite{ask_me_anything, decaNLP}. Interpreting tasks (such as event extraction) as question answering enables us to leverage zero-shot capabilities of pre-trained models. Moreover, these methods are not dependant on domain-specific features, but solely on plain text. Given the zero-shot capabilities of MRCP tasks across domains \cite{gpt3}, it is more likely that this procedure transfers across domains. 

Our proposed task has many similarities with the computational identification of \textit{framing} \cite{card-etal-2015-media} and \textit{agenda setting} \cite{tsur-etal-2015-frame,field-etal-2018-framing}, as well as with automated \textit{bias measurement} \cite{bolukbasi2016man,caliskan2017semantics}.
These various tasks all seek to identify the author's written perspective; the same topic can be portrayed differently by different authors, just as the hero or victim might be assigned differently by different authors \cite{bergstrand2018villains}. 
Our identification of the hero, villain, and victim provides yet another method to describe the particular viewpoint expressed in a particular text and to draw comparisons between these various viewpoints over large datasets.

\section{Methods}

\subsection{Labeling Character Roles}

Our approach is to frame the labeling of narrative character roles as a machine reading comprehension or closed  question answering task. We use auto-regressive language models --- i.e., we provide the question and context as prompts to a pre-trained model and decode the answer span token-by-token. We use GPT-3 \cite{gpt3}, which has proven proficiency in various question-answering tasks \cite{squad_v2}. This method allows us to directly leverage knowledge acquired in pre-training on vast amounts of text. 

Figure \ref{fig:fig_1} shows an example prompt. We directly ask, ``Who is the villain [or hero or victim] in the following text?''. That question is followed by the story text, and then the respective character role is repeated to nudge the model to generate the most likely completion of this prompt. We use the same prompt across all experiments in this study, only varying the story text. We use the 175B-parameter davinci model with default decoding parameters.\footnote{except for the state-of-the-union address speeches where we use the 13B-parameter curie model for cost reasons.}

To benchmark our new model's performance, we consider as a baseline the dictionary-based model from \cite{NFP_and_NLP}. First, they use named entity recognition (NER) to extract important entities from news articles. Second, for each entity, they use the surrounding text and its sentiment polarity and dictionary matching to decide whether an entity is a hero, villain, or victim.\footnote{We could not find the original source code for \cite{NFP_and_NLP}, so we used the implementation available at \url{https://github.com/meganzhao10/Hero-Villain-Victim}.}

\subsection{Corpora}

We apply our labeling approach to three corpora, described here.
These corpora span three domains and types of narratives: descriptions of current events, fictional stories, and political speeches.

\paragraph{Newspaper Articles.} The first domain is newspaper articles. We use a corpus of 66 newspaper articles about fracking published in the \textit{Boulder Daily Camera}, a local Colorado newspaper, from the years 2008-2013. \cite{Blair_and_McCormack} hand-code the three character roles (hero, villain, victim) in these articles.\footnote{Note that the original article also had articles from the \textit{Colorado Springs Gazette}, but we were unable to reproduce that part of the  corpus. We contacted the authors to provide the articles, but without success.} The average length of each article is 682 words.

\paragraph{Disney Movie Plots.}  The second domain is Disney Movie plots. We selected eleven Disney movies based on a ``most well-known classics'' list (see Table \ref{tab:disney_movies} below). We then downloaded the plot summary section for these movies from Wikipedia. The average plot summary length is 670 words.

\paragraph{U.S. Presidential Speeches.} Our third corpus includes presidential speeches given at the annual U.S. State of the Union Address, for the years 2001 to 2018.\footnote{These are hosted on Kaggle at \url{https://www.kaggle.com/datasets/rtatman/state-of-the-union-corpus-1989-2017}.} We split each speech into paragraphs and skip paragraphs containing fewer than 20 words. The final corpus contains  $N=1,379$ paragraphs. Each paragraph contains on average 73 words.

\section{Results}

This section presents the results, with the three empirical domains reported in turn.

\begin{table}[]
    \centering
    \footnotesize
    \begin{tabular}{l c c c}
     Character  &  Accuracy GPT-3 & Accuracy Baseline & N \\ \hline
     \rule{0pt}{2.5ex}Hero & 50\% & 15\% & 20 \\
     Victim & 90\% & 65\% & 20 \\
     Villain & 47\% & 18\% & 17 \\ \hline
     \rule{0pt}{2.5ex}All & 63\% & 33\% & 57 \\ \hline
         & 
    \end{tabular}
    \caption{Main Evaluation Results: Accuracy of GPT-3 for extracting heroes, villains and victims from The Boulder Daily Camera articles, compared to a dictionary-based baseline descried in \cite{NFP_and_NLP}. In the last column \textit{N}, we show the number of annotations for each character type present in the data.}
    \label{tab:main_results}
\end{table}

\subsection{Newspaper Articles about Fracking}

Our first  analysis applies our GPT-3 method to the collection of news articles about fracking from \cite{Blair_and_McCormack}. That paper uses the manual annotations of character roles to analyze framing differences between liberal and conservative media. Regardless of the political leaning, the media outlets in that study framed the public as the victim and the oil and gas industry as the villain. However, the role of hero differed: the liberal media outlet often presented environmental organizations as the hero, while this role is instantiated by specific actors of the oil and gas industry in the conservative outlet.

To automate the annotation process, we use the prompt shown in Figure \ref{fig:fig_1} for each article, the difference being that the text now is the article in question. In the manually annotated data, the authors only find 20 heroes, 20 victims and 17 villains, but our method produces a result for every character role in every article. For evaluation, we only consider model outputs in cases where a true gold annotation exists, and discard all other articles. 


In the gold labels from \cite{Blair_and_McCormack}, annotations are coarsened such that each specific role (e.g. hero, villain) is mapped to one of a finite set of classes: the public, the government, environmental organizations, or the oil and gas industry. A challenge in the model evaluation is that the language model is not constrained to the finite label set, so the generated text output often does not exactly match the gold labels, even when the output is semantically correct. For the purposes of evaluation, we manually map each GPT-3-generated answer to one of the four categories. The set of GPT-3 outputs and our annotated labels are shown in Appendix Table \ref{tab:appendix_table_main_results}.

  \begin{table*}[ht]
     \centering
     \footnotesize 
     \begin{tabular}{l p{1.8cm} p{5.1cm} p{4.9cm}}
     Movie & Hero & Victim & Villain \\ \hline
    \rule{0pt}{2.5ex}\textit{101 Dalmatians} & Roger Dearly & The Dalmatian Puppies & Cruella de Vil \\
    \textit{Aladdin} & Aladdin & Aladdin & Jafar \\
    \textit{Cinderella} & Cinderella & Cinderella & Lady Tremaine \\
    \textit{Alice In Wonderland} & Alice & Alice & The Queen of Hearts \\
    \textit{The Jungle Book} & Mowgli & Mowgli & Shere Khan, a man-eating Bengal tiger \\
    \textit{Sleeping Beauty} & Prince Phillip & Aurora & Maleficent \\
    \textit{The Lion King} & Simba & Mufasa & Scar \\
    \textit{Peter Pan} & Peter Pan & Wendy, John, Michael, and the Lost Boys & Captain Hook \\
    \textit{Mary Poppins} & Mary Poppins & Mr. Banks & Mr. Dawes \\
    \textit{The Little Mermaid} & Ariel & Ariel & Ursula \\
    \textit{Snow White} & Snow White & Snow White & The Queen \\ \hline
     \end{tabular}
     \caption{Results for Wikipedia plots of widely known Disney Movies}
     \label{tab:disney_movies}
 \end{table*}

The evaluation results are shown in Table \ref{tab:main_results}.  We achieve an overall accuracy of 63\%, a large improvement over the dictionary baseline from \cite{NFP_and_NLP} (33\%). While both methods provide decent results for the victims (which is usually assigned to the public), our approach achieves strong gains in detecting the heroes and villains. We observe an almost three-times improvement for both villains and heroes. 

\subsection{Disney Movie Plot Summaries}
 
Next, we provide qualitative evidence that our method also works in a second domain of popular movie plot summaries. We extract heroes, villains and victims from Wikipedia plot descriptions for widely known Disney movies. Given that these movies contain well-known heroes and villains (if not always victims), it is straight-forward to manually evaluate the quality of the extracted roles. 
 
The list of annotations for the Disney moves are reported in  Table \ref{tab:disney_movies}. Readers who are familiar with the movies can see that the method works very well in this setting. While some of these annotations are limited or arguable, none are indefensible --- there is some reasonable argument for each of these 33 annotations being correct.

\subsection{U.S. State of the Union Addresses}

In our last application, we show how the method can be used to analyze political discourse in the context of U.S. State of the Union Address speeches, where there is no labeled data, as in the fracking articles, or easily verified set of roles, as in the Disney movies. As we have no ground-truth labels, this section follows a descriptive social-science approach and includes adaptations to our previous methods to improve the legibility of the results.

As before, we apply the method to extract a victim, hero, and villain in each paragraph from the corpus of recent U.S. State of the Union Addresses. The free-form texts generated for the character roles are diverse. We have hundreds of unique answers for each role, with many singletons. To reduce the dimensionality of these outputs and make them more interpretable, we encode the phrases using S-BERT  \cite{sbert} and apply $k$-means clustering to the resulting vectors \cite{jahan-etal-2021-inducing, ash_narratives}. After manual inspection for different $k$, we select $k$=20.

We then use the partisan affiliation of the speakers to score the most Democrat-associated and most Republican-associated clusters in each character role. Formally, we compute the log odds ratio of each cluster w.r.t. the party affiliation of the president giving the speech and show the cluster with the highest and lowest odds ratio.

\begin{table*}[]
    \centering
    \scriptsize
    \begin{tabular}{l p{7.5cm} p{6cm}}
        Role & Democrats & Republicans  \\ \hline
        \rule{0pt}{2.5ex}Hero & The average family watching tonight, the average person, The average American household, The average person, The average worker, Average American. \textbf{Log Odds Ratio: -0.88} & The men and women of the 9/11 generation who have served in Afghanistan and Iraq, The United States military, The military, The veterans, The Cajun Navy volunteers, The man who lost four of his brothers at war, The troops, The troops and civilians who sacrifice every day to protect us, America's veterans . . . 
        \textbf{Log Odds Ratio: 1.0} \\
        \rule{0pt}{2.5ex}Victim & The American students, The community colleges, The American student, The person who pays for the good education., The school district, A student, The American public school system, The school, The students who are not American citizens, The students, High school students, The high school graduates in Germany, the American student, The teacher, The school system, Every high school diploma is a ticket to success. \textbf{Log Odds Ratio: -1.43} & the coalition to defeat ISIS, ISIS leader, al-Baghdadi, Assad, The UN concluded that Saddam Hussein had biological weapons sufficient to produce over 25,000 liters of anthrax, enough doses to kill several million people, The President of the Iraqi Governing Council, Safia Taleb al-Suhail, Prime Minister Allawi, Iraqi security forces, Iraqi interpreter, Iraqi Government, The Iraqi Government, The American and Iraqi surges have achieved results few of us could have imagined just one year ago. \textbf{Log Odds Ratio: 1.42} \\
        \rule{0pt}{2.5ex}Villain & The college, The teacher who comes in early because he knows she might someday cure a disease., The school administration, The educational system, The school in Dillon, South Carolina, The national competition to improve schools is the villain in this text., The school, The Education Secretary, The education reformer, The school system. \textbf{Log Odds Ratio: -1.37} & The Taliban, Islamic State, ISIS leader, al-Baghdadi, Intel, The terrorists, The unnamed terrorist group, The terrorist underworld, Al Qaida, International terrorism, Iraqi officials, Iraqi intelligence officers, The enemies of freedom, Radical Islam, Marines, Al Qaeda, Sunni extremists, Syrian occupiers, Radical Shia elements. \textbf{Log Odds Ratio: 1.81} \\ \hline
         & 
    \end{tabular}
    \caption{Heroes, Victims and Villains extracted from State of the Union speeches. Shown in this table are the entries for the cluster with the highest/lowest odds ratio for Democratic and Republican Presidents}
    \label{tab:log_odds}
\end{table*}

Table \ref{tab:log_odds} displays the clusters with the highest partisan log odds ratio by character role --- that is, the entities taking on this role more often for one or the other party. For Republican presidents (Bush and Trump), the heroes, victims, and villains in SOTU addresses are connected to the U.S. military and wars in the Middle East. Democratic speeches (by Obama) have a more populist flavor, with the average American portrayed as a hero. Intriguingly, for Democrats the villains and victims are both associated with the education system.

\section{Discussion and Future Work}

\paragraph{Task formulation.}
Perhaps the highest-priority limitation of our study is that the method will try to extract a character role from a text, when prompted, even when the role is not present. The newspaper-article evaluation metrics would be much worse if we included the articles missing a role in the test set. In the presidential speeches, in particular, we frequently found that our model assigned the same agent to all three roles --- even though villain is mutual exclusive with hero or victim --- because there was only one agent mentioned in the speech. 
There are a number of ways to address this issue. Perhaps the simplest would be to adjust the prompt to allow for a ``not applicable'' answer, or to ask a preliminary question: ``Does this text contain a [role]?'' For both of these adjustments, a few-shot approach where the model is provided with some examples could improve performance.  

\paragraph{Prompt engineering.}
Prior work has shown that prompts with subtle differences can product significantly different results \cite{holtzman-etal-2021-surface, calibrate_before_using}. Besides few-shot learning, the language-model prompting could also be adjusted to potentially improve performance. Rather than asking about the three roles in three separate prompts, the model could be asked to identify all three simultaneously, for example. The question could be asked in different ways and then the answers aggregated. One could also explore adapting the prompt to constrain the set of entities to a finite set; e.g., in the fracking articles corpus, there was a pre-specified list of four possible entities. Finally, GPT-3 has some decoding hyperparameters that could be tweaked. 

\paragraph{Limitations of large language models.}
Like other NLP models \cite{bolukbasi2016man}, large pre-trained language models can encode harmful human biases \cite{bender2021dangers}.
For example, prior work has shown that narratives generated by GPT-3 explicitely portray feminine characters as less powerful \cite{lucy-bamman-2021-gender} while also encoding implicit gender biases \cite{huang-etal-2021-uncovering-implicit}. 
GPT-3 is trained on multiple large datasets, including scraped web text, book texts, and Wikipedia articles.
Because of their size, these datasets are difficult to document; even quantifying the number of duplicate documents can be a challenge \cite{lee2021deduplicating} and even more difficult are detailed descriptions, like those called for in data documentation best practices \cite{gebru2021datasheets}.
We use GPT-3 to measure authors' framing biases, but it is simultaneously likely that biases encoded in GPT-3 influence our results in ways that are difficult to measure.

Another major limitation to the use of the GPT-3 API is the cost of OpenAI API queries. The queries for our relatively small-scale analysis of state-of-the-union address speeches cost nine dollars using the 13B-parameter GPT-3 model. Scaling this up to larger corpora of thousands or millions of documents, such as the \textit{U.S. Congressional Record} \cite{ash_narratives}, would be prohibitively expensive. Hence, using even larger pre-trained models, such as PaLM \cite{chowdhery2022palm}, is likely not cost-effective for most academic research. Exploring smaller open-source language models, such as GPT-Neo \cite{gpt-neo}, which can be implemented at scale, is a promising alternative.

Moving beyond pre-trained language models, performance and scalability could be improved through further model training. Fine-tuning GPT-3 for this task is one possibility. A less expensive option would be to use GPT-3 to create a labeled dataset, perhaps with human supervision, for training a smaller student model. That student model could be a distilled autoregressive model or an encoder model like BERT. For MRCP, BERT-like approaches work well for question-answering tasks where the answer is a span of tokens in the input text --- in our case, the character being assigned a character role. A limitation of this approach is that character roles are often implicit, rather than explicitly mentioned in the text. BERT-like models work for explicit mentions, but for implicit mentions a generative model like GPT is needed.

\paragraph{Concluding note.}

In this work, we used the zero-shot capabilities of a large-scale language model to automatically extract heroes, villains and victims from newspaper articles, movie plot summaries, and U.S. presidential speeches. Large pre-trained language models can solve machine reading comprehension tasks for the purposes of labeling short to medium-sized documents, without hand-annotated training data. This approach could be useful for countless projects in computational social science and digital humanities. 

While promising, our results are still a proof of concept. We have introduced a basic version of the method, which performs better than prior work using a dictionary baseline. But our method's sufficiency for social-science applications is not yet assured, and more work is needed to build up the method and assess its robustness in the field.



\bibliographystyle{unsrtnat}
\bibliography{references}

\appendix




\begin{table*}[]
    \centering
    \scriptsize
    \begin{tabular}{p{6.7cm} p{0.8cm} p{2.8cm} p{4cm} p{1.2cm}}
    Article title & Role & Annotated & GPT-3 answer & Equivalent \\ \hline
No. 3: Gloves come off in the fight over fracking & villain & Oil and gas industry & the oil and gas industry & 1 \\
Polis to oil and gas: please stop suing the communities I represent & hero & Fed Govt & Jared Polis & 1 \\
Anti-fracking measures win in Lafayette, Boulder, Fort Collins & victim & General or specific public & Lafayette & 1 \\
Anti-fracking measures win in Lafayette, Boulder, Fort Collins & villain & Oil and gas industry & the oil and gas industry & 1 \\
Colorado Oil and Gas Aaaociation gives \$600K to fight fracking bans on Front Range Ballots & hero & Oil and gas industry & Merrily Mazza & 0 \\
Colorado Oil and Gas Aaaociation gives \$600K to fight fracking bans on Front Range Ballots & victim & General or specific public & Lafayette Campaign for Energy Choice & 1 \\
Colorado Oil and Gas Aaaociation gives \$600K to fight fracking bans on Front Range Ballots & villain & Enviromnental Orgs & The Colorado Oil and Gas Association & 0 \\
Lafayette to draft resolutions of opposition to fracking ban, utility occupation tax & victim & General or specific public & Lafayette & 1 \\
Broomfield approves agreement with oil and gas operator & hero & Enviromnental Orgs & The City Council & 0 \\
Broomfield approves agreement with oil and gas operator & victim & General or specific public & Anthem residents & 1 \\
Broomfield approves agreement with oil and gas operator & villain & Local Govt & Sovereign & 0 \\
Ballot initiative to ban fracking in Lafayette garners enough signiatures to move forward & victim & General or specific public & Lafayette & 1 \\
Extension of Boulder County fracking moratorium deemed a win for many & victim & General or specific public & the citizens of Boulder County & 1 \\
Lafayette anti-fracking effort draws canvassers & hero & Enviromnental Orgs & Cliff Willmeng & 1 \\
Lafayette anti-fracking effort draws canvassers & victim & Oil and gas industry & Lafayette & 0 \\
Lafayette anti-fracking effort draws canvassers & villain & Enviromnental Orgs & East Boulder County United & 1 \\
Broomfield unlikely to pursue fracking moratorium & hero & Enviromnental Orgs & The City Council & 0 \\
Broomfield unlikely to pursue fracking moratorium & victim & General or specific public & the community in Broomfield & 1 \\
Broomfield unlikely to pursue fracking moratorium & villain & Oil and gas industry & The City Council & 0 \\
Boulder County Commissioner Elise Jhones to debate Gov John Hickenlooper on fracking & hero & Local Govt & Boulder County Commissioner Elise Jones & 1 \\
Boulder County Commissioner Elise Jhones to debate Gov John Hickenlooper on fracking & victim & General or specific public & Boulder County Commissioner Elise Jones & 0 \\
Boulder County Commissioner Elise Jhones to debate Gov John Hickenlooper on fracking & villain &  State Govt & Elise Jones & 1 \\
Anti-fracking group to present Lafayette with petition for ban & hero & Enviromnental Orgs & Anti-fracking group & 1 \\
Anti-fracking group to present Lafayette with petition for ban & victim & General or specific public & Lafayette & 1 \\
Anti-fracking group to present Lafayette with petition for ban & villain & Oil and gas industry & East Boulder County United & 0 \\
Lafayette to pursue moratorium on new oil and gas permits & hero & Oil and gas industry & Lafayette & 0 \\
Group urges ban on fracking Activists ask Lafayette for moratorium & hero & Enviromnental Orgs & John Chinnery & 0 \\
Group urges ban on fracking Activists ask Lafayette for moratorium & victim & General or specific public & the community of Lafayette & 1 \\
Group urges ban on fracking Activists ask Lafayette for moratorium & villain & Enviromnental Orgs & John Chinnery & 0 \\
In Erie, oil and gas companies to pay twice as much for water & hero & Enviromnental Orgs & The town last week doubled its commercial water rate . . . & 0 \\
In Erie, oil and gas companies to pay twice as much for water & victim & General or specific public & The town of Erie & 1 \\
In Erie, oil and gas companies to pay twice as much for water & villain & Oil and gas industry & The oil and gas companies & 1 \\
Anti-fracking group to provide air-monitoring equiptment to erie residents & hero & General or specific public & The anti-fracking group Erie Rising. & 0 \\
Anti-fracking group to provide air-monitoring equiptment to erie residents & hero & Enviromnental Orgs & The anti-fracking group Erie Rising. & 1 \\
Anti-fracking groups to protest Encanas drilling operation near Erie schools Saturday & hero & Enviromnental Orgs & Anti-fracking groups & 1 \\
Anti-fracking groups to protest Encanas drilling operation near Erie schools Saturday & victim & General or specific public & Erie schools & 1 \\
Anti-fracking groups to protest Encanas drilling operation near Erie schools Saturday & villain & Oil and gas industry & Encana & 1 \\
Erie mothers, children descend on Encana headquarters with anti-fracking petition & victim & General or specific public & Erie mothers and children & 1 \\
Dirty air in Erie linked to gas drilling & hero & Enviromnental Orgs & The researcher with the National Oceanic and Atmospheric Administration & 0 \\
Dirty air in Erie linked to gas drilling & victim & General or specific public & Erie residents & 1 \\
Dirty air in Erie linked to gas drilling & villain & Oil and gas industry & Industry & 1 \\
Lawsuit from COGA brings Lafayette into thick of fight over oil and gas drilling & hero & Fed Govt & Lafayette & 0 \\
Pro-fracking group alleges 'systemic failures' in Broomfield election & hero & Enviromnental Orgs & BJ Nikkel & 1 \\
Colorado Oil and Gas Association gives \$600K to fight fracking bans on Front Range ballots & victim & General or specific public & Lafayette campaign for energy choice & 1 \\
Voter frustration fills ballot with citizen initiatives in Boulder, Broomfield counties & hero & General or specific public & A frustrated voter & 1 \\
Voter frustration fills ballot with citizen initiatives in Boulder, Broomfield counties & villain & Other & - & 0 \\
Broomfield tightening requirements for oil and gas companies & villain & Oil and gas industry & The City Council & 0 \\
Coming off win at Boulder County level, anti-frackers turn focus statewide & hero & Enviromnental Orgs & Anti-frackers & 1 \\
In talk at CU-Boulder, Hickenlooper says he is 'constantly attacked now for being in the pocket of oil and gas' & hero & Oil and gas industry & Governor John Hickenlooper & 0 \\
Broomfield postpones hearing on North Park fracking application & hero & Enviromnental Orgs & Jackie Houle & 1 \\
Broomfield postpones hearing on North Park fracking application & victim & General or specific public & Concerned residents of Broomfield & 1 \\
'Bucket Brigade': Anti-fracking citizen effort to monitor the air in Erie & victim & General or specific public & Erie residents & 1 \\
'Bucket Brigade': Anti-fracking citizen effort to monitor the air in Erie & villain & Oil and gas industry & Global Community Monitor & 0 \\
Hundreds gather to protest Encana Corp.'s fracking operation in Erie & victim & General or specific public & The community & 1 \\
Hundreds gather to protest Encana Corp.'s fracking operation in Erie & villain & Oil and gas industry & Encana Corp. & 1 \\
Erie eyes agreements with oil and gas operators & villain & Oil and gas industry & Erie & 0 \\
Fracking discussion packs Erie Town Hall, no action taken on moratorium & victim & General or specific public & The community of Erie & 1 \\ \hline
    \end{tabular}
    \caption{Article title, annotated label from \cite{Blair_and_McCormack}, the GPT-3 output and the author's determination whether the generated output is equivalent to the manual annotation.}
    \label{tab:appendix_table_main_results}
\end{table*}
\end{document}